\pdfoutput=1

\documentclass[11pt]{article}

\usepackage[final]{acl}

\usepackage{times}
\usepackage{latexsym}
\usepackage{amsmath}
\usepackage{enumitem}
\usepackage{graphicx}
\usepackage{booktabs}
\usepackage{tabularx} 
\usepackage{makecell}
\graphicspath{ {./images/} }

\usepackage[T1]{fontenc}

\usepackage[utf8]{inputenc}

\usepackage{microtype}

\usepackage{inconsolata}

\usepackage{graphicx}

\title{MIS-LSTM: Multichannel Image–Sequence LSTM for Sleep Quality and Stress Prediction}

\author{
    Seongwan Park \\ Sungkyunkwan University \\ Republic of Korea \\ \texttt{waniboyy@gmail.com}
    \And  Jieun Woo \\ Sungkyunkwan University \\ Republic of Korea \\ \texttt{wjieun@g.skku.edu}
    \And  Siheon Yang \\ Yeungnam University \\ Republic of Korea \\ \texttt{tlgjs7006@yu.ac.kr}
}

\begin{document}
\maketitle

\begin{abstract}
\small
This paper presents MIS-LSTM, a hybrid framework that joins CNN encoders with an LSTM sequence model for sleep quality and stress prediction at the day level from multimodal lifelog data.  Continuous sensor streams are first partitioned into N-hour blocks and rendered as multi-channel images, while sparse discrete events are encoded with a dedicated 1D-CNN.  A Convolutional Block Attention Module fuses the two modalities into refined block embeddings, which an LSTM then aggregates to capture long-range temporal dependencies.  To further boost robustness, we introduce UALRE, an uncertainty-aware ensemble that overrides low-confidence majority votes with high-confidence individual predictions.  Experiments on the 2025 ETRI Lifelog Challenge dataset show that Our base MIS-LSTM achieves Macro-F1 0.615; with the UALRE ensemble, the score improves to 0.647, outperforming strong LSTM, 1D-CNN, and CNN baselines. Ablations confirm (i) the superiority of multi-channel over stacked-vertical imaging, (ii) the benefit of a 4-hour block granularity, and (iii) the efficacy of modality-specific discrete encoding. 
\end{abstract}


\section{Introduction}

Poor sleep quality and high stress levels are related to serious health consequences, including elevated risks of cardiovascular disease, depression, impaired cognition, and weakened immune function. Chronic sleep deprivation also disrupts emotional regulation, creating a vicious cycle between stress and sleep problems \cite{scataglini2023, zhou2022}. These issues have driven growing interest in lifelog data from wearables and smartphones as a means to continuously monitor daily behaviors and infer sleep and mental well-being. Lifelog sensor streams (e.g., activity, heart rate) provide rich information reflecting an individual’s daily lifestyle, which can yield insights into sleep patterns and stress in naturalistic settings \cite{song2023}. Advanced wearable sensors combined with deep learning models enable unobtrusive health monitoring from such data \cite{chriskos2021}, and have already shown promise in automatic sleep-stage recognition \cite{supratak2017}.

\begin{figure}[!htbp]
  \centering
  \includegraphics[width=0.48\textwidth]{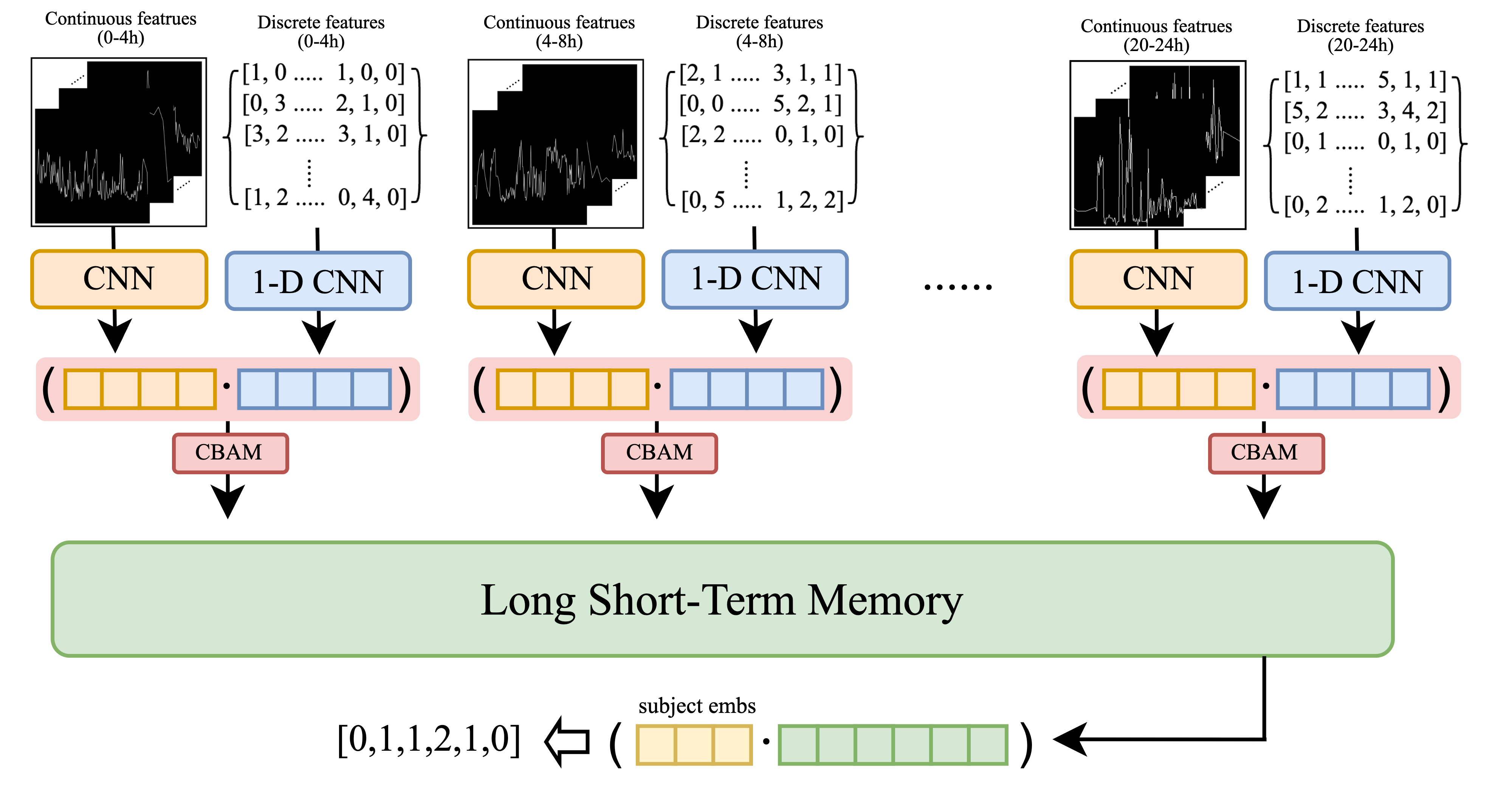}
  \caption{Overview of a Multichannel Image–Sequence LSTM (MIS-LSTM) framework (4-hour block intervals).}
  \label{fig:framework}
\end{figure}

Recent advances in time-series modeling fall into two complementary paradigms. On one hand, sequential architectures—such as Long Short-Term Memory networks (LSTMs) and Time-Series Transformers (TSTs) \cite{zerveas2021}, which—being inspired by the Transformer architecture \cite{vaswani2017}—capture long-range sequences. However, their emphasis on global sequence modeling often comes at the expense of fine-grained, non-linear motifs that manifest over short intervals. On the other hand, image-based approaches embed fixed-length windows of the series as two-dimensional “images” and apply convolutional neural networks (CNNs). Through sliding convolutional filters, CNNs are well-suited to detecting localized temporal patterns—such as periodic cycles, abrupt transitions, and peaks—by convolving over adjacent time steps \cite{fawaz2020}. Moreover, by allocating each continuous variable to its own channel, these models learn cross-channel interactions via 3D kernel (C×H×W), preserving cross-feature interactions while encoding each modality efficiently. However, despite their powerful local feature extraction, CNNs’ fixed receptive fields and the implicit loss of exact temporal ordering limit their ability to reason over long‐range temporal dependencies.

To integrate the complementary strengths of convolutional and LSTM architectures while addressing their respective shortcomings, we introduce MIS-LSTM: Multichannel Image–Sequence LSTM, a hybrid CNN–LSTM architecture. In our day-level sleep quality and stress prediction framework, raw sensor streams are first divided into fixed-length, \textbf{N-hour blocks}. Within each block, we distinguish between two feature types:

\begin{itemize}
    \item Continuous features (e.g., heart rate, distance traveled) are real-valued and dense, capturing fine-grained physiological or motion signals 
    \item Discrete features (e.g., activity category, smartphone on/off status) are integer-valued and sparse, encoding categorical or event-driven states.
\end{itemize}

Because these modalities exhibit different properties, jointly convolving them can impede learning; instead, each is processed through a dedicated CNN encoder optimized for its characteristics. The block-wise feature maps produced by these encoders are then concatenated and passed through a Convolutional Block Attention Module (CBAM) \cite{woo2018} to emphasize salient patterns across feature maps, yielding a refined N-hour representation. Finally, the sequence of these block representations serves as input to a LSTM, which captures global, day-level temporal dependencies. This two-stage pipeline (i) excels at extracting localized, nonlinear motifs via CNNs and (ii) integrates these local features over longer horizons through the LSTM, leading to superior performance on lifelog-based time-series tasks. Our method is summarized in Figure \ref{fig:framework}.

Furthermore, we propose a novel ensemble strategy that improves robustness by leveraging prediction confidence. Multiple model variants are trained, and for each sample, we examine the logit margin (difference between the top two or three class scores) as a confidence measure. For ambiguous cases (low margin), standard majority voting is used, whereas for certain cases (high margin) where the top-performing model strongly disagrees with the ensemble majority, we selectively override the majority vote with the confident prediction. This adaptive voting mechanism systematically resolves conflicting votes by privileging the most confident model, effectively correcting cases where the majority might be mistaken.

Experimental results show that on a public lifelog dataset from the 2025 challenge (predicting daily sleep quality (Q1–Q3) and sleep health metrics (S1–S3)), we demonstrate that the proposed framework significantly outperforms baselines. In particular, it yields substantially higher Macro-F1 scores than: (a) an raw LSTM baseline, (b) a raw CNN and only use 1D-CNN, (c) our model variant in which the dedicated discrete feature branch is omitted. We also conduct ablation studies to validate the benefit of multi-channel image encoding over alternative representations (tested on N-hour blocks with N in 2,4,6),  and ensemble refinement on performance.

The contributions of this paper are as follows:
\begin{enumerate}
    \item Hybrid Sequential–Image Framework: We introduce a two-stage CNN– LSTM architecture for time-series prediction that encodes time-series into N-hour block representations for local pattern extraction via CNN, and integrates these representations across N-hour blocks with an LSTM to capture long-range temporal dependencies.
    \item Modality-Specific Feature Modeling: We design a hybrid processing pipeline that preserves feature characteristics by handling continuous sensor streams through multi-channel CNN encoding and discrete, sparse signals through a dedicated 1D-CNN branch, yielding superior performance compared to merging all features in a single representation.
    \item Uncertainty-Aware Logit-Refined Ensemble (UALRE): We propose a novel ensemble strategy that computes logit margins as a confidence signal, applying standard majority voting for low-confidence samples and selectively overriding the vote when a highly confident model conflicts with the ensemble, thereby improving robustness over traditional soft or hard voting schemes.
\end{enumerate}
\section{Related Work}

\begin{table*}[!htbp]
\scriptsize

\caption{Overview of multimodal lifelog sensor streams, including frequency (Hz.), data types, and brief descriptions of each feature.}
\label{tab:data_items}

\centering
\begin{tabularx}{\textwidth}{lllX}
\toprule
\textbf{Items} & \textbf{Freq.} & \textbf{Data type} & \textbf{Note} \\
\midrule
mACStatus     & 1/60   & integer & 0: No, 1: Charging \\
mActivity     & 1/60   & integer & \makecell[l]{0: in\_vehicle, 1: on\_bicycle, 2: on\_foot, 3: still, 4: unknown, 5: tilting, 7: walking, 8: running} \\
mAmbience     & 1/120  & object  & List of ambient sound labels and their probabilities \\
mBle          & 1/600  & object  & List of Bluetooth device address, device\_class, and RSSI \\
mGps          & 1/60   & object  & List of (altitude, latitude, longitude, speed) \\
mLight        & 1/600  & float   & Ambient light in lx unit \\
mScreenStatus & 1/60   & integer & 0: No, 1: Using screen \\
mUsageStats   & 1/600  & object  & List of app names and their usage times (ms) \\
mWifi         & 1/600  & object  & List of base station ID (BSSID) and RSSI \\
wHr           & 1/60   & object  & List of heart rate recordings \\
wLight        & 1/600  & float   & Ambient light in lx unit \\
wPedo         & 1/60   & \makecell[l]{float; float; float; \\ integer; float} & \makecell[l]{Number of calories; Distance in meters; Speed in km/h unit; Number of steps; Step frequency in a minute} \\
\bottomrule
\end{tabularx}
\end{table*}

\begin{table*}[!htbp]
\scriptsize

\caption{Definitions and value encodings for the six daily sleep and stress metrics used as prediction targets.}
\label{tab:metrics}

\centering
\begin{tabularx}{\textwidth}{lll}
\toprule
\textbf{Metric} & \textbf{Explanation} & \textbf{Values} \\
\midrule
Q1 & Overall sleep quality as perceived immediately after waking up & 0: Below individual average, 1: Above individual average \\
Q2 & Physical fatigue level just before sleep & 0: High level of fatigue, 1: Low level of fatigue \\
Q3 & Stress level experienced just before sleep & 0: High level of stress, 1: Low level of stress \\
S1 & Adherence to sleep guidelines for total sleep time (TST) & 0: Not recommended, 1: May be appropriate, 2: Recommended \\
S2 & Adherence to sleep guidelines for sleep efficiency (SE) & 0: Inappropriate, 1: Recommended \\
S3 & Adherence to sleep guidelines for sleep onset latency (SOL) & 0: Inappropriate, 1: Recommended \\
\bottomrule
\end{tabularx}
\end{table*}

\subsection{Lifelog-Based Sleep and Stress Prediction}
The problem of predicting sleep quality and stress from personal lifelogs has gained traction recently, particularly through competitive research challenges. In the 2024 ICTC “Predicting Sleep Quality and Emotional States” multi-label challenge \cite{oh2024}, several teams proposed novel deep learning solutions for lifelog prediction. For example, Kim et al. \cite{kim2024_tram} proposed TraM, which leverages a Time Series Transformer (TST) for labels with strong temporal dynamics and a Machine Learning Ensemble for labels requiring aggregated daily statistics. Similarly, Kim et al. \cite{kim2024_lifelog} introduced a LSTM–based framework that applies data augmentation—specifically, time-shifting and noise injection—to generate multiple lifelog variations, followed by an ensemble of these variants to obtain the final prediction. While these methods demonstrate strong performance and effectively model the characteristics of time series data, they inherently struggle to capture fine-grained local temporal patterns that convolutional operations excel at extracting. In contrast, our approach employs a convolutional neural network (CNN), enabling its convolutional kernels to explicitly learn local temporal features prior to higher-level sequence modeling.

\subsection{Multivariate Time-Series Data as Images}
Several recent works have reformulated multivariate time series data as images for CNN‐based prediction. PixleepFlow \cite{na2025} constructs each sensor feature as a 2D matrix—mapping time to the x-axis and feature magnitude to the y-axis—and then vertically stacks these matrices into a single‐channel (grayscale) image for sleep quality and stress prediction. In contrast, Oh et al. \cite{oh2022} assign each feature its own image channel, thereby leveraging the spatial inductive bias of CNNs more effectively than the single‐channel approach of PixleepFlow. Although these image‐based methods excel at capturing local temporal motifs, they inherently do not encode long‐range dependencies as sequential models do. Our MIS-LSTM framework bridges this gap by first extracting fine‐grained feature representations with a multichannel CNN and then integrating them over extended horizons using a LSTM.

\section{Dataset and Preprocessing}

\subsection{2025 ETRI Lifelog Analytics Challenge Dataset}
We evaluate our approach on the publicly released dataset from the 2025 ETRI Lifelog Analytics Challenge, which focuses on predicting daily sleep quality and stress-related metrics from multimodal sensor data. The dataset consists of smartphone and smartwatch lifelog data collected from multiple participants over a continuous period (several weeks). Each day of data is labeled with six target metrics: three subjective self-reports and three objective sleep measurements. The data were recorded via participants’ Android smartphones and wearable devices with sampling intervals ranging from 1 to 10 minutes \cite{oh2024}. Due to varying usage patterns, the raw data contains occasional missing values and noise (e.g., periods when a device was not worn or was charging). We describe the input features and output metrics below.

The detailed structure of individual sensor data (input) items is summarized in Table \ref{tab:data_items}. Each data item is stored as an individual data file and is provided along with the participant ID and timestamp. In the 2025 Challenge, six labels (Q1–Q3, S1–S3) are provided following the format of prior years, as detailed in Table \ref{tab:metrics}. Q1–Q3 are binary labels derived from daily questionnaires (above vs. below each individual’s mean), while S1–S3 are adherence-based sleep metrics (ternary for total sleep time; binary for sleep efficiency and sleep onset latency), computed against National Sleep Foundation guidelines \cite{nsf2025}.

Although the challenge prescribes 450 days for training and 250 days for held-out testing (with test labels withheld), we approximate an end-to-end evaluation by applying an 80/20 stratified split—at the subject level—on the publicly available data, ensuring proportional representation of each participant in both subsets. All model development and hyperparameter tuning are performed on the training partition, with final performance reported on the validation partition. We frame the task as multi-label classification, predicting all six daily metrics simultaneously, and evaluate performance via the weighted sum of their Macro-F1 scores.

\begin{table}[!tbp]
\scriptsize

\caption{Discrete and continuous sensor features, their per‐interval aggregation descriptions, and resulting feature counts.}
\label{tab:features}

\centering
\begin{tabularx}{0.485\textwidth}{lXc}
\toprule
\textbf{Feature name} & \textbf{Description} & \textbf{Features \#} \\
\midrule
\multicolumn{3}{c}{\textbf{Discrete}} \\
\midrule
mActivity      & Total counts of activity codes within each 30-minute interval: vehicle (0), bicycle (1), still (3), walking (7). & 4 \\
mAmbience      & Total counts of ambient context levels within each 30-minute interval: low (0), medium (1), high (2). & 3 \\
mScreenStatus  & Total screen-on duration within each 30-minute interval. & 1 \\
mACStatus      & Total charging duration within each 30-minute interval. & 1 \\
\midrule
\multicolumn{3}{c}{\textbf{Continuous}} \\
\midrule
mBle           & Max Bluetooth RSSI per 1-minute interval & 1 \\
mGps           & Travel distance per 1-minute interval & 1 \\
mUseageStats   & Sum of app usage time per 1-minute interval & 1 \\
mWifi          & Max Wi-Fi RSSI and device count per 1-minute interval & 2 \\
wHr            & Mean heart rate per 1-minute interval & 1 \\
wLight         & Ambient light level per 1-minute interval & 1 \\
\bottomrule
\end{tabularx}

\end{table}

\subsection{Data Preprocessing}
To prepare the raw multimodal lifelog streams for our MIS-LSTM framework, we employ a modality‐aware preprocessing pipeline that jointly addresses feature relevance, temporal regularization, noise filtering, and normalization. First, we restrict the input space to those features shown to be predictive in prior work \cite{kim2024_lifelog, na2025}, thereby reducing redundancy and computational overhead (Table \ref{tab:features}). Continuous, real‐valued features are resampled via linear interpolation onto a uniform one‐minute windows (1,440 timesteps per day), producing dense, evenly spaced series suitable for convolutional encoding. In contrast, discrete, integer‐valued event features are inherently sparse and are therefore aggregated into ten‐minute windows (144 timesteps per day). Within each window, we compute summary statistics (event counts, durations) to capture essential patterns without introducing excessive sparsity. To mitigate excessive impact of outlier, values that lie outside are clipped to valid bounds or removed entirely. Finally, each channel is standardized to zero mean and unit variance. The outcome of this pipeline is two modality‐specific matrices—one for the high‐resolution continuous grid and one for the discretized event sequence—that are then processed by dedicated encoder branches (a multi‐channel 2D‐CNN and a lightweight 1D‐CNN, respectively), enabling the network to learn modality‐specific representations.
\section{PROPOSED METHODOLOGY}
Our framework consists of two sequential stages: 
\begin{enumerate}
    \item \textbf{CNN-based sensor feature encoding}, Continuous features—sampled at one-minute intervals—and discrete features—sampled at ten-minute intervals—are partitioned into N-hour blocks and processed by modality-specific CNN encoders, respectively, generating a sequence of block-level feature maps for each day
    \item \textbf{Multichannel Image Sequence LSTM}, The resulting sequence of block-level feature maps for both continuous and discrete modalities is then fed into a unified LSTM, which jointly attends to local patterns within each block and global dependencies across blocks to capture comprehensive temporal relationships.
\end{enumerate}

\subsection{Continuous Feature Encoding with Multi-Channel N-hour Block Image}
As described in Section III, we employ modality-specific encodings for discrete and continuous features. Continuous features are first segmented into N-hour blocks, each of which is rendered as a two-dimensional, grayscale image (where the x-axis denotes time and the y-axis represents the corresponding feature value at each time point). This process produces K separate images—one for each feature. However, since each N-hour block must represent all K feature maps simultaneously, we merge these K images into a single, unified representation. 

\begin{figure}[t]
  \centering
  \includegraphics[width=0.45\textwidth]{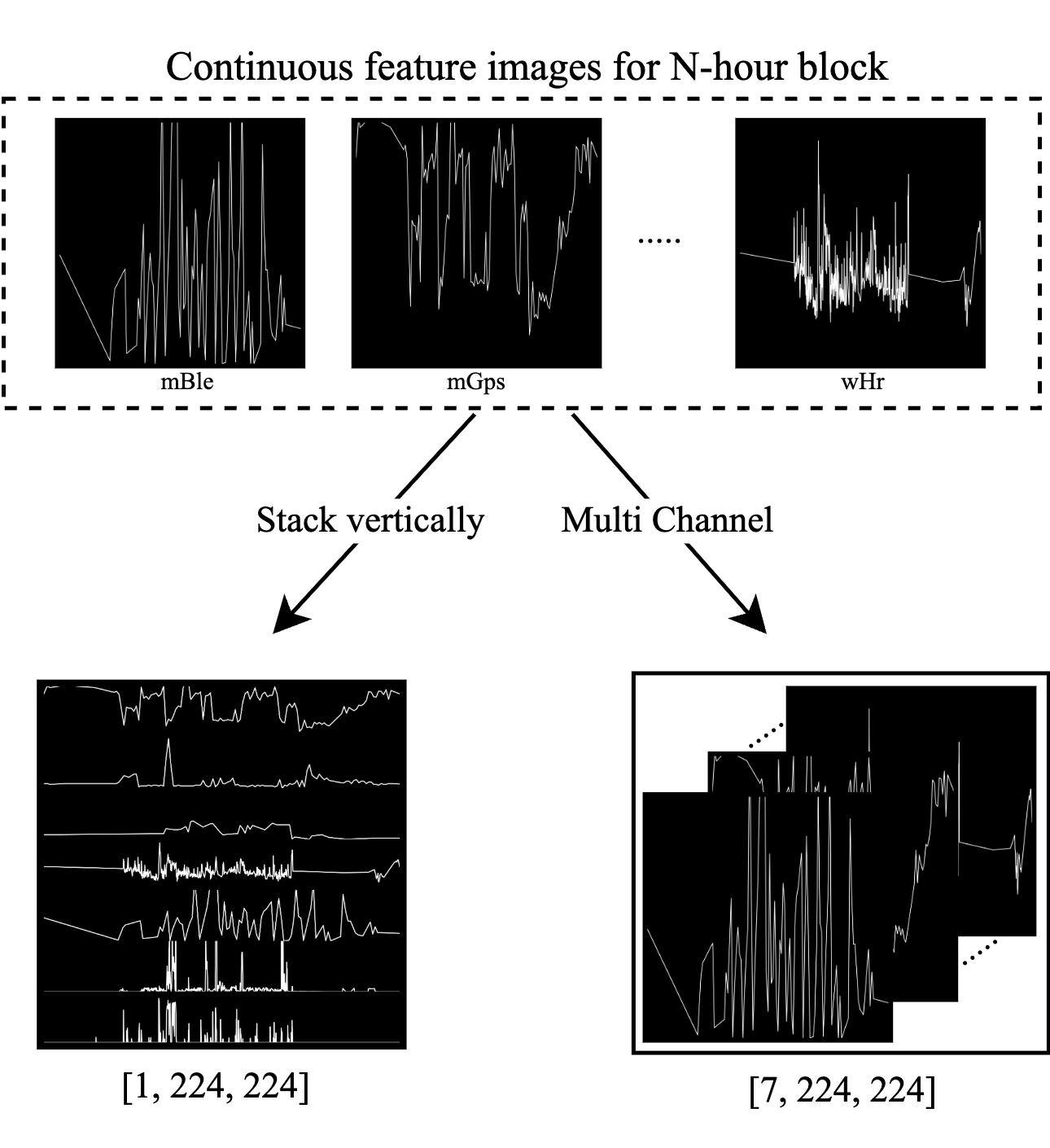}
  \caption{Illustration of two strategies for representing multivariate time-series images as a unified image. Left: Stacked-Vertical (single-channel), Right: Stacked-Channel (multi-channel)}
  \label{fig:channel}
\end{figure}

Conventional time-series CNNs employ a “stacked vertical” encoding, in which each feature’s image is concatenated along the vertical axis into a single unified image (Figure \ref{fig:channel}, left). However, we found that using such a single-channel (gray scale) image leads to inferior performance, as the CNN’s convolutional filters struggle to separate features effectively when they are overlaid in one channel. Instead, we adopt a “stacked channel” encoding, (Figure \ref{fig:channel}, right) where each sensor feature becomes one channel in a multi-channel image for the block—analogous to the R, G, and B channels of a color image. This multi-channel method enables the network to apply feature-specific kernels and then integrate them, yielding superior learning of complex patterns. Finally, integrated N-hour block image is fed into our ResNet-based CNN encoder to produce an N-hour embedding for downstream sequence modeling.

\subsection{Discrete Feature Encoding with 1D-CNN}
\begin{figure}[!tbp]
  \centering
  \includegraphics[width=0.48\textwidth]{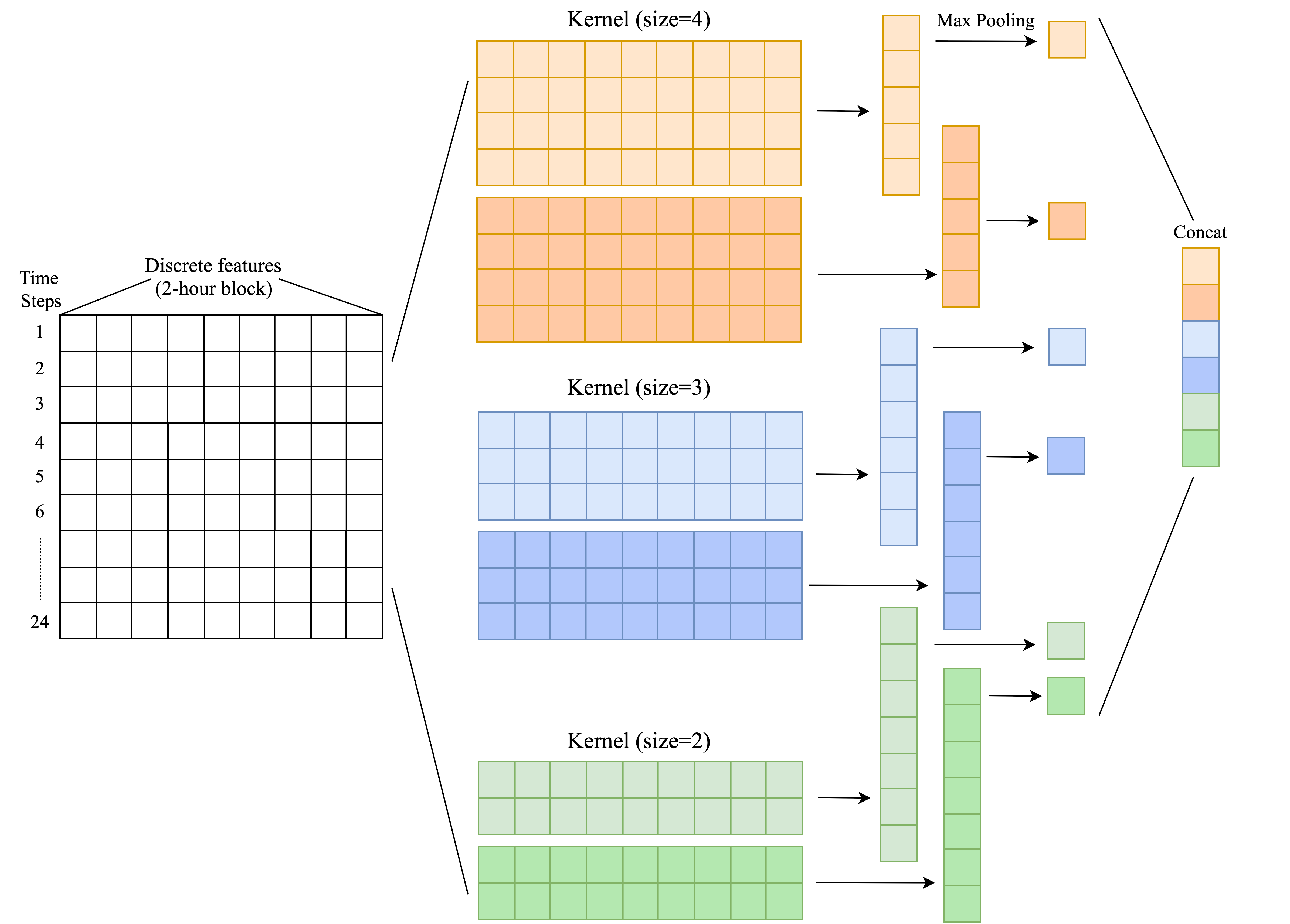}
  \caption{Discrete feature encoding via 1D-CNN. The input is a matrix whose rows correspond to time steps and columns to discrete features. Multiple one-dimensional convolutional filters of varying kernel sizes are applied along the temporal axis; each resulting feature map is max pooled and then concatenated to form the unified discrete embedding.}
  \label{fig:encoding}
\end{figure}

Unlike continuous features, discrete features indicate the occurrence or category of an event at specific time points rather than varying continuously in magnitude. Moreover, because discrete signals tend to be relatively sparse, their advantages when represented as two-dimensional images are limited. Encoding both continuous and discrete modalities with the same 2D-CNN can therefore impair overall performance. To address this, we employ a dedicated one-dimensional convolutional network (1D-CNN) for discrete data. By leveraging this modality‐specific feature modeling, we obtained feature maps that faithfully preserved the intrinsic characteristics of each data modality (Figure \ref{fig:encoding}).

\subsection{LSTM for Image-Encoded Sequences}
The embeddings produced by the CNN encoder for each N-hour block effectively capture local patterns within blocks but are limited in modeling the temporal continuity and long-range dependencies across blocks (i.e across hours). To address this limitation and characterize day-level representation, we introduce an LSTM (Long Short-Term Memory) layer. We first concatenate the N-hour block embeddings derived separately from continuous and discrete features, then apply a Convolutional Block Attention Module (CBAM) to emphasize salient channels and spatial regions. The refined sequence of block embeddings is subsequently fed into the LSTM, enabling the model to learn both inter-block temporal continuity and long-term dependencies. Finally, the LSTM’s final hidden state is passed to a classification layer to predict the six target metrics (Q1–Q3 and S1–S3). Additionally, because the 2025 Lifelog Challenge dataset comprises only ten participants, modeling individual-specific characteristics is highly advantageous. Accordingly, we introduce a learnable subject embedding for each participant and concatenate it with the LSTM’s final hidden state, thereby enabling the network to incorporate and learn identity information alongside the extracted features. This three-stage CNN–CBAM–LSTM pipeline synergistically combines fine-grained intra-block feature extraction with inter- block sequential modeling, substantially improving daily sleep quality and stress prediction performance.

\subsection{Uncertainty-Aware Logit-Refined Ensemble (UALRE)}
Based on the predictions produced by MIS-LSTM, we conducted an additional ensemble phase.  Conventional ensembles \cite{lakshminarayanan2017} typically rely on one of two strategies: Soft voting – averaging either the raw logits or their soft-maxed probabilities across models and then taking the arg max. Hard voting – ignoring the logits entirely and selecting the class chosen by the majority of models. Because both strategies assign equal weight to every model output, performance can deteriorate when a few high-quality models are outnumbered by inferior ones. To address this drawback, we introduce the Uncertainty-Aware Logit-Refined Ensemble (UALRE):

\begin{table*}[t]
\centering
\caption{Macro-F1 scores for each target metric, comparing LSTM, 1D-CNN, CNN, and our MIS-LSTM.}
\label{tab:main_result}
\begin{tabular}{lccccccc}
\toprule
\textbf{Model} & \textbf{Q1} & \textbf{Q2} & \textbf{Q3} & \textbf{S1} & \textbf{S2} & \textbf{S3} & \textbf{Avg} \\
\midrule
\textbf{LSTM}                        & 0.630 & 0.581 & 0.561 & 0.436 & 0.632 & 0.651 & 0.576 \\
\textbf{1D-CNN}                      & 0.592 & 0.554 & 0.542 & 0.468 & 0.602 & 0.599 & 0.559 \\
\textbf{CNN}                         & 0.614 & 0.567 & 0.576 & 0.414 & 0.458 & 0.649 & 0.578 \\
\textbf{MIS-LSTM}                    & \textbf{0.625} & \textbf{0.626} & \textbf{0.618} & \textbf{0.486} & \textbf{0.650} & \textbf{0.682} & \textbf{0.615} \\
\textbf{MIS-LSTM (w/o discrete)}     & 0.615 & 0.574 & 0.621 & 0.449 & 0.629 & 0.677 & 0.594 \\
\bottomrule
\end{tabular}
\end{table*}

\begin{itemize}
    \item \textbf{Step 1 (confidence filtering)}: For the best individual model, we treat samples with a large logit margin (high inter-class gap) as “confident.” These predictions are accepted without modification 
    \item \textbf{Step 2 (selective hard voting)}: For the remaining “uncertain” samples(low inter-class gap) we apply hard voting, but we restrict the ballot to predictions from other models only on the samples they deem confident (i.e., high inter-class gap). This ensures that even low-performing models contribute only when they are internally certain.
\end{itemize}

By coupling confidence filtering with selective hard voting, UALRE preserves the strengths of the most reliable model while leveraging complementary information from the ensemble only where it is trustworthy, thereby mitigating the equal-weight limitation of traditional soft- and hard-voting schemes.

\section{Experiment}

\subsection{Experimental Setup \& Baselines}
For discrete features, we employ a one‐dimensional CNN in PyTorch, instantiating 16 filters for each kernel size in {3, 4, 5, 6} and aggregating their outputs via max-over-time pooling. Continuous features are encoded with a SEResNeXt101\_32×4d backbone (timm library) \cite{xie2017}, which we train from scratch—eschewing ImageNet pretraining—and modify its first convolution to accept seven input channels (one per continuous feature) instead of three. We fuse the concatenated discrete and continuous block embeddings (N-hour blocks) and refine them with a Convolutional Block Attention Module (CBAM). To capture global temporal dependencies across N-hour blocks, a two-layer LSTM (model dimension = 256), also implemented in PyTorch.

Optimization proceeds with AdamW (learning rate 3e-5, mini-batch size = 16). Because the label distribution is markedly skewed, we used Focal Loss instead of cross-entropy. Training spans 200 epochs; to mitigate overfitting on the limited dataset, we select the checkpoint achieving the highest validation macro-F1 rather than the lowest validation loss, and report results using that model.

To assess the efficacy of our framework, we compare against the following baselines:
\begin{itemize}
    \item LSTM: All features sampled at 10-minute intervals are fed into a standard LSTM. The hidden size and number of layers match those used in our proposed model.
    \item 1D-CNN: All features sampled at 10-minute intervals are processed by a one-dimensional CNN, employing identical kernel sizes and filter counts in our proposed model and does not employ N-hour block segmentation or downstream sequence modeling.
    \item CNN: All features sampled at 1-minute intervals are converted into multi-channel images and passed through a SEResNeXt101\_32×4d backbone. This variant does not employ N-hour block segmentation nor the downstream LSTM and uses only a CNN.
\end{itemize}

\begin{table}[t]
\footnotesize
\centering
\caption{Macro-F1 comparison of three ensemble strategies.}
\label{tab:ensemble_result}
\begin{tabular}{lc}
\toprule
\textbf{Ensemble method} & \textbf{F1 score (macro)} \\
\midrule
\textbf{soft-vote} & 0.621 \\
\textbf{hard-vote} & 0.614 \\
\textbf{UALRE (ours)} & \textbf{0.647} \\
\bottomrule
\end{tabular}
\end{table}

\begin{table}[t]
\scriptsize
\centering
\caption{Macro-F1 scores for discrete feature encoding using multi‐channel versus stacked‐vertical representations at varying N-hour block lengths.}
\label{tab:channel_result}
\begin{tabular}{lccc}
\toprule
& \textbf{\makecell[c]{2hours \\ (12images/day)}} & \textbf{\makecell[c]{4hours \\ (6images/day)}} & \textbf{\makecell[c]{6hours \\ (4images/day)}} \\
\midrule
\textbf{stacked-vertical} & 0.581 & 0.591 & 0.588 \\
\textbf{multi-channel} & 0.593 & \textbf{0.615} & 0.601 \\
\bottomrule
\end{tabular}
\end{table}

\subsection{Main Result}
Table \ref{tab:main_result} summarizes the Macro-F1 scores for each target metric and the overall average across baselines and our MIS- LSTM. Our model achieves the highest average score (0.615),
outperforming the strongest baseline (CNN). These results demonstrate that (i) modality-specific CNN encoders effective-ly capture both continuous and discrete signals, and (ii) the subsequent LSTM integration robustly models day-level sequential dependencies. In particular, the largest improvements on Q2 and S3 suggest that our hybrid architecture excels at both fine-grained emotional state detection and sleep health assessment, validating the benefits of multi-channel imaging.

Additionally, to assess the impact of our Modality-Specific Feature Modeling for discrete data, we conduct an ablation study on a variant (“MIS-LSTM w/o discrete”) that instead treats discrete signals as additional channels in the same 2D-CNN used for continuous features, rather than routing them through a dedicated 1D-CNN branch. This ablation achieves an average Macro-F1 of 0.594 yet still outperforms the other baselines. The resulting performance gap confirms that processing sparse, event-driven signals via a 1D-CNN branch provides a meaningful boost in capturing discrete feature representation.

Finally, we apply our UALRE to the model outputs from Tables \ref{tab:main_result} and \ref{tab:channel_result}. As Table \ref{tab:ensemble_result} shows, UALRE achieves a Macro‐F1 of 0.647, outperforming standard soft voting (0.621) and hard voting (0.614). This demonstrates that selectively overriding low‐confidence majority decisions with high‐confidence predictions yields more robust performance than traditional voting schemes.

\subsection{Effect of N-hour Block Length}
To assess the effect of N-hour block length, we experimented with N = 2, 4, and 6 hours—yielding 12, 6, and 4 blocks per day, respectively—and report the resulting Macro-F1 scores in Table \ref{tab:channel_result}. We also compared two continuous-feature encoding strategies: multi-channel versus stacked-vertical. Across all block lengths, the multi-channel approach consistently outperformed the stacked-vertical baseline, suggesting that treating each feature as an independent channel reduces inter-feature interference and more effectively captures complex local patterns. Examining block length, N=4 hours attained the highest score (0.615), which we attribute to a trade-off between global context and local motif learning: shorter blocks fragment long-range dependencies, whereas longer blocks dilute fine-grained temporal features, with 4-hour segmentation striking the optimal balance.
\section{Conclusion}

We have presented MIS-LSTM, a novel hybrid architecture that couples multi-channel CNN encoders for both continuous and discrete lifelog features with an LSTM backbone to jointly capture fine-grained temporal motifs and long-range dependencies. By integrating a Convolutional Block Attention Module to fuse modality-specific embeddings and introducing UALRE—an uncertainty-aware logit-refined ensemble—we achieves the best performance among evaluated methods on the 2025 ETRI Lifelog Challenge. Our extensive ablations confirm the efficacy of multi-channel over stacked-vertical imaging, the optimality of a 4-hour segmentation, and the benefit of a dedicated 1D-CNN branch for sparse discrete events. Beyond raw performance, MIS-LSTM’s block-level interpretability, personalized subject embeddings, and modality-specific design position it as a practical foundation for real-time sleep and stress monitoring. We anticipate that this framework will advance personalized digital health solutions by enabling early detection of anomalies and delivering actionable insights into daily behavior patterns and stress monitoring. We anticipate that this framework will advance personalized digital health solutions by enabling early detection of anomalies and delivering actionable insights into daily behavior patterns.

\bibliography{custom}
\nocite{*}

\end{document}